\documentclass[11pt]{article}
\usepackage{url}
\usepackage{times}

\newtheorem{example}{Example}[section]

\newtheorem{definition}{Definition} [section]

\newcommand{\I}[1]{\relax\ifmmode\mbox{\it#1}\else{\it#1}\fi}

\newcommand{\tbl}{\hspace*{12mm}}
\newcommand{\no}{not\,}

\newcommand{\s}{\mbox{\,}}
\newcommand{\IF}{\mbox{:-}}
\newcommand{\DE}{\mbox{:$>$}}
\newcommand{\DA}{\mbox{:$<$}}

\begin{document}
\begin{sloppypar}

\begin{center}
\huge
\noindent
{\bf Towards Active Logic Programming
\footnote{Note of the author: This paper was presented at the
2nd International Workshop on Component-based Software Development in Computational Logic (COCL 1999), 
and appeared in the online Proceedings (A. Brogi and P.M. Hill eds., \url{http://www.di.unipi.it/~brogi/ResearchActivity/COCL99/proceedings/index.html}). In this paper, the Agent-Oriented logic programming language DALI \cite{jelia02,jelia04} was first introduced.}}\\
\medskip\medskip
\large
{\bf Stefania Costantini}\\
\normalsize
{\it Dipartimento di Matematica Pura e Applicata\\
Universita' degli Studi di L'Aquila}\\
via Vetoio Loc. Coppito, I-67100 L'Aquila (Italy)\\
stefcost@univaq.it
\end{center}

\begin{abstract}
In this paper we present the new logic programming language DALI, aimed at 
defining agents and agent systems.  
A main design objective for DALI has been that of
introducing in a declarative fashion all the essential features,
while keeping the language
as close as possible to the syntax and semantics of the plain Horn--clause language. 
Special atoms and rules have been introduced, for representing: external events, 
to which the agent is able to respond (reactivity); actions 
(reactivity and proactivity); internal events
(previous conclusions which can trigger further activity);
past and present events (to be aware of what has happened).
An extended resolution is provided, so that a DALI agent is able
to answer queries like in the plain Horn--clause language,
but is also able to cope with the different kinds of events,
and exhibit a (rational) reactive and proactive behaviour.
\end{abstract}

\section{Introduction}
\label{intro }
In this paper we address the issue of defining a logic programming language 
for reactive and proactive agent systems, with a clear procedural and declarative semantics.
The motivation of this paper is that, while it is 
quite straightforward to build logical agents which are in some way rational,
it  is much more difficult to build logical agents 
that interact with the environment, and perform actions 
either on their own initiative (i.e. they are {\em proactive}) or in response 
to events which occur externally (i.e. they are {\em reactive}).

A lot of work has been done in order to equip logical agents with 
more and more sophisticated forms of rationality.
Logical agents can represent (and communicate) their own as well as 
other agents' thinking processes,
and are able to build arguments by means of their own inference rules
 where:  either each agent has a logic associated with it, \cite{giunchiglia:ag}, 
 \cite{PSJ:ag},
with rules of inference which relate the different 
logics of the different agents;
or, each agent is
defined together with its specific inference rules 
and communication modalities \cite{CDL:meta92}.

Logical agents are able to combine belief updating, 
goal updating and practical reasoning 
in a BDI fashion (``Belief, Desires, Intentionality'') \cite{rao:91}, \cite{rao:95}.
Logical agents are able to interact with an environment on which
they have limited \cite{aix:eijk} or no \cite{aix:elio} expertise.
Logical agents can reason about actions (\cite{gelfond:azioni} and subsequent extensions)
and can manipulate not simply logical formulas, but complex data types, and are able to act 
in accordance with a specific, declarative policy \cite{vs:agrep}.
Many of these approaches are related to logic programming, 
either syntactically or
semantically, others are based on theorem--proving or 
transition systems as an operational semantics.

Reactive and proactive agents typically achieve their functionality 
by means of condition--action rules, where the condition
constitutes a stimulus that causes the agent to perform some kind of action as response.
The approach of  \cite{kowalski:rag} 
is aimed at combining rationality and reactivity in logic: it proposes that
reactivity is achieved, in a rational agent, by modeling reactive rules as 
integrity constraints in the proof procedure.
By combining the approaches of  \cite{kowalski:rag} and \cite{CDL:meta92}, 
the approach of \cite{pier:jelia98}
obtains rational reactive agents by treating communication primitives as abducibles.
There are, however, few approaches in logic programming for representing 
not only the actual behavior of {\it one} agent in a detailed way,
but also the behavior of {\it complex} multi--agent systems, with 
few notable exceptions. A relevant approach is the concurrent logic 
programming language ConGolog \cite{lesperance}, \cite{shapiro}, based on situation calculus, where properties of multi--agent systems can be formally proved. 
At present, ConGolog has been less useful as an implementation language, although further 
developments are under way. 
Another one
is the Constraint Logic Programming language CaseLP, which can also be used
in the context of a Multi-Agent-System specification methodology
based on linear logic \cite{martelliMAS}, \cite{martelliSE}.

In this paper we focus on representing reactivity and proactivity in logic programming 
in the simplest possible way, with the objective of
introducing in a declarative fashion all the essential features,
while keeping the language
as close as possible to the syntax and semantics of logic programming 
\cite{lloyd:foundations}.
The resulting language (that we call DALI) should then be easy to understand, 
easy to implement, and easy to use. 
To this aim, we introduce: 
\begin{description}
\item
distinct atoms to represent {\em external events}; whenever
an event takes place (it can be an action or a message from another agent,
or something which changes in the external world), the corresponding
event atom becomes true;
\item
distinct atoms to represent {\em internal events}; an internal event
corresponds to a conclusion reached by the agent, which can trigger some
kind of activity in the agent itself;
\item
distinct atoms to reason about {\em past} events, and
{\em present} events, the latter being events which have already taken place,
to which the agent has however not yet responded;
\item
distinct atoms to represent {\em actions} that can be performed
by the agent; 
\item
{\em reactive rules} that have an event (either external or internal)
as conclusion; a reactive rule can be understood as the agent's
{\em reaction} to the event;
\item
{\em action rules} that have an action 
as conclusion; an action rule can be understood as 
defining the {\em preconditions} for an action to be performed;
\item
{\em active rules} that have actions in their body, so as to model
proactive behavior, where an agent performs an action with the 
aim of reaching some kind of objective, and
reactive behavior where an agent acts
in response to stimuli, since a reactive rule can be active.
\end{description}

Procedural semantics is based on an 
easy--to--implement, extended resolution.
Declarative semantics of agents and agent systems is based on ``snapshots''
of the current stage of the computation, which consist 
in the Least Herbrand Models on a suitably modified versions of
the logic programs representing agents. Snapshots can then
be organized so as to represent ``evolutions'' of the agent system.

In order to keep the discussion as simple as possible,
we take as basic language the plain propositional Horn-clause language,
and we just consider ``events'' and ``actions'' without even trying any
finer distinction. We are conscious of the fact that, when introducing variables and
time, there will be many related issues to be coped with. Also
for lack of space, we leave these topics out of the scope of the present paper.
In perspective, however, the practical logic language DALI
will have all the useful features of real languages,
though still remaining (in our opinion) as simple and clean as to
be easily integrated, if needed, with most of the previously mentioned
approaches.

\section{Reactive Logic Programming }
\label{reactive}
\subsection{External Events}
\label{events}
In order to illustrate the extension we propose to the Horn--clause language, 
it is useful to take into consideration
 about how rules are understood in logic programming.
Consider  for instance the following logic program:
\begin{eqnarray*}
p. & &\\
p\s & \IF & r,s
\end{eqnarray*}
The second rule is by no means necessary to prove the truth of  $p$, 
that is an immediate consequence of the first 
rule (which is a fact). In this case, clearly  truth of $p$  
is not conditioned by truth of $r, s$. 
It is important to notice however that
the second rule is still a candidate for resolution of goal $?-p$, 
and it will be finally selected by any fair
interpreter, and will be applied by the immediate consequence operator $T_{P}$. 
Then, since we already know that $p$ is true, 
we can interpret the resolvent $?-r,s$ as a {\em reaction} 
of the interpreter in {\em response} to the {\em stimulus} $?-p$.

Let us  now assume that $p$ is an {\em external event}, i.e.
something which occur outside (and independently of) the program
that we are considering. 
External events, or stimuli, could be messages, or actions 
(performed by some other entity) 
which affect the agent, or observations,
depending of the environment where the program
is put at work. At present, we do not
distinguish between these cases. 

Similarly to most logical approaches to agents, we leave to the language
developer to decide how the agents are made conscious that an event took
place, and how to suitably represent an event to a given agent 
(for instance, if you call me on the phone, I see this action the other way round,
i.e. in my view I am being called). 

We simply assume that, as soon as this event happens,
 $p$ somehow assumes value true for every agent Ag which is able to observe this event.
 Then, any rule defining $p$ in the program of agent Ag, 
 can be interpreted as a stimulus--response rule, i.e.
a rule that determines some actions to be executed in response to the stimulus $p$.

Enhancing logic programs with some kind of modules of agents 
(or, more generally, of multiple theories) is a
well studied topic. Following \cite{CDL:meta92}, we assume we have a form 
of modularization available, that allows a program to be divided
into separate modules, or agents, each one endowed with its own name
(this without affecting the standard semantics). 
In the following, when referring to agent $Ag$, 
we will implicitly refer to the logic program defining $Ag$.

Then, we associate to $Ag$ a set of distinct
atoms 
\[E_{Ag} = \{E_{1},\ldots,E_{s}\}, \mbox{\,\,\,}s \geq 0\] 
representing external events. We call each of the $E_{j }$'s an {\em event atom}.
According to  \cite{CDL:meta92}, whenever event  $E_{j}$ takes place, 
atom  $E_{j }$ becomes true for all the other agents. We assume that events
which have taken place are recorded in a set $EV \subseteq E_{Ag}$

For each $E_{j }$, $Ag$ may possibly provide {\bf only one} rule of the form: 
\[ E_{j} \IF  R_{j,1},\ldots,R_{j,q}. \tbl q\geq 1\]
where the $R_{j,i}$'s can be understood as {\em responses} to the stimulus $E_{j }$.

We call this rule a {\em reactive rule}. Notice that having a single reactive rule
for each event is not really a restriction, since the $R_{j,i}$'s may 
have multiple definitions.

We assume that events are ''consumed'' by reactive rules. I.e., the application
of the reactive rule causes $E_{j}$ to be removed from $EV$

We state 
the limitation that the  $E_{j }$'s do not appear in the body of other rules. 
Then, when $E_{j }$ happens, we are in a situation fairly analogous to that of the above
example, since we have a true atom, and a rule which is 
by no means aimed at proving this fact, but that, instead, specifies activities
(internal inferences and/or actions)
to undertake as response. Then, {\em atoms in the body should not be
intended as determining the truth of $E_{j }$}, but, vice versa,
as {\em predicates that should be proved whenever $E_{j }$ becomes true}. 
In a sense,
it is like if the opposite side of the implication is being used. However,
notice that we do not say that truth of $E_{j }$ {\em implies} truth
of $R_{j,1},\ldots,R_{j,q}$, but that {\bf on truth} of $E_{j }$,
truth of $R_{j,1},\ldots,R_{j,q}$ should be checked. 

To make this interpretation more intuitive, and in order to make it recognizable that 
rules with conclusion $E_{j }$ are stimulus--response rules, 
a slightly different notation may be used, 
for instance by replacing the token  ``$ \IF $'' with ``$ \DE $'',
Then the syntax of a reactive rule becomes: 
\[ E_{j} \DE  R_{j,1},\ldots,R_{j,q}. \tbl q\geq 1\]

For the sake of readability, given predicate $p$, the fact that $p\in E_{Ag}$,
i.e. that $p$ represents an external event,
is stressed by an explicit subscript, by denoting $p$ as $p_E$.

\begin{example}
If an event determines some behavior, then stimulus--response
rules are adequate, like for instance
\begin{eqnarray*}
Ag & &\\
\I{rains$_E$} &  \DE  & \I{open\_umbrella}.\\
\I{open\_umbrella} & \IF & \I{have\_umbrella}.
\end{eqnarray*}
where $rains \in E_{Ag}$. The rule says that in case it rains, the agent
opens an umbrella. Then, on truth of event 
rains, we check the truth of (i.e. try to prove) the related condition 
\I{open\_umbrella}. This is done as usual in logic programming,
i.e. using the 
second rule, and try proving its condition \I{have\_umbrella}.
\end{example}

Notice that, if there are several atoms in the conditions of a reactive rule,
their order may be relevant.
\begin{example}
Consider the following variation of previous example.
\begin{eqnarray*}
Ag & &\\
\I{rains$_E$} &  \DE  & \I{open\_umbrella}, \I{decide\_what\_to\_do}.
\end{eqnarray*}
In this case, it is better to open the umbrella {\em before}
deciding what to do next (which for instance could be going home, or 
instead entering a shop).
\end{example}
\subsection{Actions}
\label{actions}
Since $Ag$ is reactive and proactive, we also associate to
$Ag$  the set of distinct atoms 
\[A_{Ag} = \{A_{1},\ldots,A_{q}\}, \tbl q \geq 0\] 
which represents the actions that Ag is able to perform. 
We call each of the $A_{j }$'s an {\em action atom}.

Actions can be performed in response to external events, 
but also on the agent's own initiative
(proactivity).
The $A_{i}$'s can appear in the body of rules, and may 
have an explicit definition, i.e. 
there can be (optionally)  {\bf only one} rule with head  $A_{i}$,
that we call {\em action rule}. 
In this case, the body of this  rule expresses preconditions 
for the action to be performed. 
Action rules are the same as 
ordinary rules, and in fact are treated by DALI resolution in
exactly the same way.
However, in order to make it visible that, conceptually,
action rules express preconditions for actions,
we add again some syntactic sugar 
so as to distinguish {\em action rules} from the others.
An action rule will have in particular the form
\[ A_{i} \DA  C_{i,1},\ldots,C_{i,s}\tbl s\geq 1\]
It is left to the implementation
that, whenever  $A_{i}$ succeeds, on the one hand 
atom  $A_{i}$ becomes true (as an external event) to all the other agents
(according to  \cite{CDL:meta92}), and, on the other hand, 
the corresponding action is performed in practice, 
in case $Ag$ actually interacts with an environment.
If there are several rules with the same action in their body,
then this action can be potentially performed several times.

Given predicate $p$, the fact that $p\in A_{Ag}$,
i.e. that $p$ represents an action,
is stressed by an explicit subscript, by denoting $p$ as $p_A$.

\begin{example}
\begin{eqnarray*}
Ag & &\\
\I{danger$_E$} &  \DE  & \I{ask\_for\_help}.\\
\I{ask\_for\_help} &  \IF  & \I{call\_police$_A$}.\\
\I{ask\_for\_help} &  \IF  & \I{scream$_A$}.\\
\I{call\_police$_A$} &  \DA  & \I{have\_a\_phone}.
\end{eqnarray*}

In this case, we have the external event \I{danger}\,$ \in E_{Ag}$, to which the agent
reacts with \I{ask\_for\_help} (which is an ordinary predicate). 
This implies possibly performing
one or both of the two actions \I{scream$_A$} or \I{call\_police$_A$}
where \I{scream$_A$} $\in A_{Ag}$, \I{call\_police$_A$} $\in A_{Ag}$.
The latter one however can be performed upon precondition \I{have\_a\_phone},
as specified by the corresponding action rule.
\end{example}

Notice that it is somewhat arbitrary to decide which are the 
predicates that are to be inserted into the set $A_{Ag}$ of actions.
In principle, the distinction between actions and ordinary predicates
is that actions affect the environment, and/or are observable from
the other agents.

From the declarative point of view, action subgoals without a corresponding
action rule always succeed, while action subgoals with an action rule
succeed or fail (in which case the action is not performed) according
to the standard procedural semantics of the Horn clause language.
In practice however, an action may be unsuccessful in the sense that,
for some reason, it is not possible to achieve the intended effect on
the external environment. For instance, referring to the above example, the 
action \I{call\_police$_A$} might in practice be prevented by the phone being out
of order. In some cases, the agent itself might cope with this kind of failures,
by following some protocol in the interaction with other agents.
In other cases, the implementation should make the agent aware of the failure
of an action. A possibility is that of specially generated external events:
in this way the agent might provide, whenever necessary, suitable ''recovery''
event rules.

\subsection{Internal  Events: Reacting to Conclusions}
\label{intevents}
Assume that, whenever some
conclusion is reached, this triggers some kind of activity
(internal reaction) in the agent. 
This is possible by defining another set of distinct
atoms, $I_{Ag}$, which contains those predicates which
are to be considered as events, and then can possibly appear as the
conclusion of {\bf only one} reactive rule.

This kind of predicates will be called {\em internal events}. 
From the procedural point of view, we assume the conclusion \I{happy}, 
obtained by means of the ordinary
rule, is  recorded in a set $IV \subseteq I_{Ag}$, and then reconsidered later
to trigger the reactive rule. Similarly to external events, the internal events
which are ''consumed'' by reactive rules, are removed from $IV$.

Given predicate $p$, the fact that $p \in I_{Ag}$,
i.e. that $p$ represents an internal event,
is stressed by an explicit subscript, by denoting $p$ as $p_IE$.

\begin{example}
Assume  that we want to express the fact that Henry,
 when happy,
merrily sings a song. Then, Henry's theory might become:

\medskip
\begin{eqnarray*}
Henry & & \\
\I{happy} &  \IF   & \I{sunny\_day}.\\
\I{happy$_{IE}$} &  \DE  & \I{sing\_a\_song}.
\end{eqnarray*}
where $\I{sing\_a\_song} \in A_{Henry}$. For defining the reactive rule, 
we let $\I{happy} \in I_{Henry}$, i.e. we state that
the internal conclusion \I{happy} may be interpreted as an  event, and determine an action. 
\end{example}

Internal events can play an important role whenever the agent makes some
kind of planning for achieving its goals. In fact, by means of reactive rules
related to internal events, plans can be ``tuned'' according to the subgoals
that have been actually achieved.

Also, internal events may help simulate a sort of ''consciousness'' in the 
agent, which is able to recognize, reason about and react to its own
conclusions. To this aim, subgoals corresponding to predicates belonging to 
$I_{Ag}$ should be automatically attempted from time to time. Referring to
the above example, the subgoal \I{happy} should be attempted every now and then,
so as to apply the reactive rule in case of success. Which ones to attempt,
and how frequently, can be left to the implementation, possibly guided by
suitable directives.  

An action atom may be considered as an
internal event, and therefore the agent
can ``react'' to its own actions.

\begin{example}
Assume  now that Anne is invited by her friends to go out with them.
She checks for the possibility of going by car, and if the car is not
available, then she takes the bus. If she take the car, she calls on
her friend Susan in order to ask her to join. 
This is represented by the following rules.

\begin{eqnarray*}
Anne & & \\
\I{invitation$_E$} &  \DE   & \I{go\_out}.\\
\I{go\_out} &  \IF  & \I{go\_by\_car$_A$}.\\
\I{go\_out} &  \IF  & \I{take\_the\_bus$_A$}.\\
\I{go\_by\_car$_A$} &  \DA  & \I{car\_available}.\\
\I{go\_by\_car$_{IE}$} &  \DE  & \I{ask\_susan\_to\_join}.
\end{eqnarray*}

\end{example}

\subsection{Drawing Conclusion from Past Events}
\label{pastevents}

It may be useful to allow event atoms in the body of rules. 
In fact, it may be the case that there is a conclusion
to draw, depending on what has happened before. 

Then, we suppose that external and internal events are recorded.
In particular, for every event atom $E \in E_{Ag} \cup I_{Ag}$, we add another
distinct atom $E^P$ (where $P$ stands for $past$), 
meaning that event $E$ has happened
in the past. We call $EP_{Ag}$ the set of these atoms.

\begin{example}
For instance, if we want to express that
George is happy if his girlfriend has called, then we need a rule 
such as:
\begin{eqnarray*}
George & & \\
\I{happy} &  \IF   & \I{girlfriend\_call}^P.
\end{eqnarray*}
where \I{girlfriend\_call} $\in E_{Henry}$. 
\end{example}

Notice that this is a source of nonmonotonicity
in the observable behavior of the agent. 
In fact, query \I{?-happy} to agent \I{Henry} may initially
fail, and may later succeed when the event \I{girlfriend\_call} will have happened. 
In our interpretation, an event becomes a past event whenever the
agent has reacted to it, i.e. as soon as the corresponding reactive
rule has been applied. 

We do not mean past events ad an ''ad hoc'' way of handling time.
Rather, we see the set of past events as a sort of ``state'' of the agent,
although it is important to notice that the facts that are recorded
are either external events which have happened, or conclusions 
that have been proved, or actions that have been performed.
The set of past events constitutes a repository for lemmas,
and therefore is no harm from the declarative point of view.
From the practical point of view, it is a sort of ''memory'',
useful to the agent to enforce a coherence in its interactions 
with the external environment.
Notice however that the set of past events is by
no means similar to the memory of imperative languages,
where one can record
arbitrary statements.

\subsection{Drawing Conclusion from Present Events}
\label{nowevents}

It can also be useful to reason about an external event which
''has already taken place'', and to which the agent has not yet reacted.
I.e., we would like atoms to be allowed into the body of rules, 
which
correspond
to events that are already available to the agent, although 
the corresponding reactive rule has not been applied yet.
This because we want to distinguish between {\em reasoning} about events,
and {\em reacting} to events.

In particular, for every event atom $E$ we add another
distinct atom $E^N$ (where $N$ stands for $now$), 
meaning that event $E$ has happened
but has not been considered yet.
We call $EN_{Ag}$ the set of these atoms.

\begin{example}
In this example Mary is awakened by the alarm clock.
Then, she realizes it is time to stand up,
and also switches the alarm clock off.
\begin{eqnarray*}
Mary & & \\
\I{my\_god\_its\_late} &  \IF   & \I{alarm\_clock\_rings}^N.\\
\I{my\_god\_its\_late$_{IE}$} &  \DE   & \I{stand\_up}.\\
\I{alarm\_clock\_rings$_E$}&  \DE   & \I{switch\_it\_off}.
\end{eqnarray*}
In this case,  \I{alarm\_clock\_rings} is an external event, while 
\I{my\_god\_its\_late} is an agent's conclusion, which is interpreted as
an internal event. Notice that the ring of the alarm clock is first
reasoned about, by means of the special predicate \I{alarm\_clock\_rings}$^N$,
which allows a conclusion to be reached, and then gives rise to a reaction.
Precisely, $\I{alarm\_clock\_rings}\in E_{Mary}$, 
$\I{my\_god\_its\_late} \in I_{Mary}$, 
$\I{switch\_it\_off} \in A_{Mary}$, and 
$\I{alarm\_clock\_rings}^N \in EN_{Mary}$
\end{example}

Present events may help reasoning about the {\em effects} of actions.
\section{Procedural Semantics}
\label{semantics}
What we need for building a DALI interpreter is the possibility
 of monitoring external and internal events, so as to actually
respond to stimuli. We propose to do that by means of an extension 
to SLD--resolution.

We assume to associate the following sets
to the goal which is being processed by a DALI interpreter:
\begin{itemize}
\item
the set $EV \subseteq E_{Ag}$ 
of the external events
that are available to the agent (stimuli to which the agent can possibly respond);
\item
the set $IV\subseteq I_{Ag}$ of internal events which have 
been proved up to now (internal stimuli to which the agent can possibly respond);
\item
the set $PV \subseteq EP_{Ag}$ of past events (both internal
and external).
\end{itemize}

The procedural behaviour
of a DALI agent may consist of the following activities. 
First, trying to answer a query (like in plain Horn--clause language).
Second, responding to either external or internal events.
Third, trying to prove a goal corresponding to an internal
event (as suggested before, these goals should be attempted from 
time to time).
These different kinds of activities are in principle independent of each other,
and should be somehow interleaved: for instance, while trying to answer a 
query, an external or internal event may occur, to which the agent should in
the meanwhile respond.  

Therefore, a goal in DALI is a {\em disjunction} $G^1;G^2;\ldots;G^n$
of {\em component goals}. Every $G^k, k \leq n$ is a goal as usually defined
in the Horn--clause language, i.e. a conjunction. The meaning is that
the computation fails only if all disjuncts fail. 

The suggested strategy
for proving a goal is the interleaving, i.e. the interpreter at each step
should be able to pick up a subgoal from any of the $G^k$'s, or to add
a new $G^j$, that in particular will be an external or internal event.
The resolution strategy adopted in a DALI
implementation will specify
how to perform the interleaving among the component goals.
The precise definition of a resolution strategy is however out of the 
scope of this paper. Instead, we concentrate in defining which are
the resolution steps that the interpreter can perform.

In fact, below is the formal definition of the extended resolution.
Notice that a resolution step will also
update (whenever necessry)
the sets $EV$, $IV$ and $PV$. The definition is composed of six different
cases. In cases (i)-(iii) the selected atom belongs to
one of the existing component goals.
In cases (iv)-(vi) instead, an atom is chosen from the set
of external (resp. internal) events, and is added to the given goal,
as a new component goal.
At each stage of the inference process, more than one case will
be in general applicable. The resolution strategy 
will state in which order the different
cases should be applied, and how often to consider the different
classes of events.

\begin{definition}[DALI Resolution]
\label{dalires}
Given a logic program defining agent $Ag$ 
(with associated sets of external events $E_{Ag}$,
internal events $I_{Ag}$,
actions $A_{Ag}$, past events $EP_{Ag}$, present events $EN_{Ag}$), 
given sets $EV \subseteq E_{Ag}$, $IV \subseteq I_{Ag}$ and $PV \subseteq EP_{Ag}$, 
and goal $G$ of the form
\[G^1;\ldots;G^n\]
where each $G^k, k \leq n$ is of the form:
\[?- {Q^k}_{1},\ldots,{Q^k}_{m}.\] 
DALI resolution derives a new goal $G'$
and new sets $EV'$, $IV'$ and $PV'$ by means of one of the following steps.
\begin{itemize}
\item[(i)]
Select atom ${Q^k}_{i} \in G^k$ and a corresponding defining clause $C$, 
and apply SLD--resolution with ${Q^k}_{i}$ 
as the selected atom and $C$ as the input clause,
thus obtaining $G'$. Let $EV' = IV$, $IV' = IV$ and $PV' = PV$.
\item[(ii)]
Select atom ${Q^k}_{i}\in G^k$ where ${Q^k}_{i}\in A_{Ag}$, 
without defining clauses, and 
derive the new goal $G'$ 
where the component goal ${G^k}$ 
is replaced by \[?-{Q^k}_{1},\ldots,{Q^k}_{i-1}, {Q^k}_{i+1},\ldots,{Q^k}_{m}.\] 
Let $EV' = EV$, $PV' = PV$ and if ${Q^k}_{i}\in I_{Ag}$ then 
$IV' = IV \cup \{{Q^k}_{i}\}$, else $IV' = IV$.
\item[(iii)]
Select atom ${Q^k}_{i}\in EN_{Ag}\cap EV$, and 
derive the new goal $G'$ 
where the component goal ${G^k}$ 
is replaced by \[?-{Q^k}_{1},\ldots,{Q^k}_{i-1}, {Q^k}_{i+1},\ldots,{Q^k}_{m}.\] 
Let $EV' = EV$, $IV' = IV$ and $PV' = PV$.
\item[(iv)]
Choose atom $E_{j}\in EV$, and
join it to the given goal, thus
deriving the new goal $G'$ with the new component goal $G^{n+1} = E_{j}$
Let $EV' = EV \setminus \{E_{j}\}$, $IV' = IV$ and $PV' = PV \cup\{E_{j}\}$.
\item[(v)]
Choose atom $I_{j}\in IV$, and
join it to the given goal, thus 
deriving the new goal $G'$ with the new component goal $G^{n+1} = I_{j}$
Let $EV' = EV$, $IV' = IV \setminus \{I_{j}\}$ and $PV' = PV \cup\{I_{j}\}$.
\item[(vi)]
Choose atom $A \in I_{Ag}$, and
join it to the given goal, thus 
deriving the new goal $G'$ with the new component goal $G^{n+1} = A$.
\end{itemize}
\end{definition}

Given a goal $G$, according to the above 
definition, DALI resolution can do one of the following. 

\begin{description}
\item
(i) Pick up a subgoal from any component goal $G^k$, and proceed with the refutation 
as usual. This case is also adequate if the subgoal is an action atom with a 
defining clause, or an event atom with a corresponding reactive clause.
\item
(ii) Perform an action without preconditions, 
which means select
an action atom without a defining clause,
and just remove it
from the goal. I.e., an action without preconditions always ``succeeds''.
It is left to the implementation perform the action, and to make the action atom true 
(in the form of an external event)
to all the other agents. If this action is among the internal events,
then it must be included into the set $IV$ since it has taken place,
and then it can possibly trigger some internal reaction.
\item
(iii) Consider an atom $E^N$ corresponding to an event $E \in EV$,
i.e corresponding to an external event which has happened,
but that has not been considered yet. This atom
can simply be removed from the goal, i.e. ``succeeds'', while $EV$ remains unchanged,
since the agent has still to react to $E$. Intuitively, the agent 
asks itself whether $E$ has happened, so as to use this knowledge in its
reasoning. Later, it will possibly react to $E$.
\item
(iv)--(v) React to any of the external or internal events, 
by inserting the corresponding new component goal in the overall goal.
The event has to be removed from
the set of the events which are still to be considered, and inserted 
into the set of past events.
\item
(vi) Insert a new component goal which corresponds to an internal event. In this
way, in case this component goal succeeds, it becomes possible to ''react'' to
this conclusion (intuitively, the agent asks itself whether $A$ holds,
so as to act consequently if this is the case). 

\end{description}

It is left to the resolution strategy which case to apply at each step
(if there are several possibilities).
I.e., it is left to the resolution strategy to choose 
how often to consider external and internal events, 
and in which order, and how to organize the interleaving among
the component goals.

What is not considered in the previous definition is the fact that,
whenever a subgoal corresponding to an internal event is proved, it
should be inserted into the set $IV$. This is for the sake of simplicity,
but notice that the definition can be made precise by introducing
{\em nested} refutation for these atoms. A {\em nested subgoal} is a subgoal
of the form $(A)_A$. 
When resolving $A$, it would become $(B_1,\ldots B_n)_A$
and so on. Any subgoal corresponding to an internal event should be
a nested subgoal.
On obtaining the empty nested subgoal $()_A$, 
i.e. on proving $A$, we would let $IV' = IV \cup \{A\}$.

Notice that $EV$ represents some kind of {\em input channel} for the agent.
As a further extension, we could have several ``channels'',
according to some classification of different events (e.g. messages
or observations or other interactions with the external environment).
Also, the resolution strategy could take as input some kind of 
{\em declaration} about the priority for selecting among different possibilities.
\section{Declarative Semantics}
\label{declarativeS}
Work is under way for a complete definition of declarative semantics, able to consider 
in general terms the evolutions of multi--agent systems. At present, we are able to specify
the declarative semantics of an agent given a certain initial situation, i.e.,
in a sense, a ``snapshot'' of the agent's behavior. We are also able to sketch
how to describe the evolution of a set of interacting agents, but we have
only an initial idea on how to study properties of the system as a whole.

In the following, we first specify semantics of a single agent,
and then introduce a concept of evolution of an agent system.

The first objective is to declaratively model reactive rules 
for external and internal events.
Consider for a moment the plain Horn--clause language, and the following program:
\begin{eqnarray*}
p. & & \\
p\s &  \IF  & q.\\
q. & &
\end{eqnarray*}
Its least Herbrand model is $\{ q, p \}$, like for the following slightly modified version:
\begin{eqnarray*}
p. & &\\
p\s &  \IF  & p,q.\\
q.
\end{eqnarray*}
Since $p$ is true by means of a unit clause, 
the second rule for $p$ does not change the meaning
of the program, since it differs from the previous version
only in that there is $p$ itself in the body. 

This is exactly the trick that we will use for our reactive Horn--clause programs. 
Precisely,
given (external or internal) event $E$, we assume to transform  rule: 
\[ E \DE  R_{1},\ldots,R_{q}. \]
into the new rule:
\[ E \IF  E,R_{1},\ldots,R_{q}. \]
Syntactically, we have added $E\s$ itself in the body of its own rule. 
The meaning is that, since there
is no other rule in Ag defining $E$, 
then the conditions of this rule may become true only if the 
truth of  $E\s$ comes from some other agent, i.e. if event  $E\s$ has happened. 
We also add rule
\[ E^P \IF  E,R_{1},\ldots,R_{q}. \]
which models the fact that as soon as the reactive rule for $E\s$ is applied,
a corresponding past event ${E^P}\s$ is generated.

The second objective is to declaratively model actions, without or with
an action clause.  
The point is, an action atom should become true (given its preconditions, 
if any) whenever the action is actually performed in some rule.
Consider another simple program written in the plain Horn--clause language:
\begin{eqnarray*}
p.& &\\
p\s &  \IF  & b,a.\\
b. & &
\end{eqnarray*}
Its least Herbrand model is $\{ p, b \}$, since both $p$ and $b$ are given as facts. 
If  we modify the program as follows:
\begin{eqnarray*}
p. & &\\
p\s &  \IF  & b,a.\\
b. & &\\
a\s &  \IF  & p,b.
\end{eqnarray*}
its least model is  $\{ p, b, a \}$. 
Assuming that $p$ is an event atom and $a$ is  an action atom with no
defining clause, this modification 
ensures that the action atom $a$ becomes true whenever the action is actually performed,
i.e. if the clause defining $p$ is
applied, and its condition $b$ is true. 
Similarly, let us assume that $a$ has a defining clause, like in the program:
\begin{eqnarray*}
p.& &\\
p\s &  \IF  & b,a.\\
b. & &\\
a &  \IF  & c.\\
c.
\end{eqnarray*}
Its least Herbrand model is $\{ p, b, a \}$, since $p$, $b$ and $c$ are given as facts. 
We modify the program as follows:
\begin{eqnarray*}
p. & &\\
p\s &  \IF  & b,a.\\
b. & &\\
a\s &  \IF  & c,p,b.
\end{eqnarray*}
Its least model is still  $\{ p, b, a \}$, but, 
interpreting $a\s$ as an action atom, we state 
then $a$ can be derived only if the corresponding action is actually performed
in the rule defining $p$. 
More generally, for every action $A$, with action rule 
\[ A \DA  C_{1},\ldots,C_{s}.\tbl s \geq 1\]
and for every other clause where $A$ appears in the conditions, of the form
\[ B \IF  D_{1},\ldots,D_{h}, A_{1},\ldots,A_{s}. \tbl h\geq 1, s\geq 1\]
with $A \in \{A_{1},\ldots,A_{s}\}$, we assume to add the new rule
\[ A \IF  B, D_{1},\ldots,D_{h},C_{1},\ldots,C_{s}.\] 
If $A$ has no defining clause, we instead add clause:
\[ A \IF  B, D_{1},\ldots,D_{h}.\] 
The meaning is that  $A\s$ is been performed (if its preconditions
$C_{1},\ldots,C_{s} \s$ hold) by the clause with head $B\s$, provided
that the conditions $D_{1},\ldots,D_{h}$ in the body of 
this clause are true.

Finally, for every
external or internal or past event $E\s$ that 
we want to be true at the beginning, we add corresponding unit clauses
\begin{eqnarray*}
E.\s\s\s\s & &\\
E^N. &  &
\end{eqnarray*}

We define the ``snapshot'' declarative semantics of a program $P$ 
written in our enhanced Horn--clause language DALI, 
as the standard declarative semantics of a program $P'$, 
obtained from $P$ by means of the modifications
specified above.
Work is under way for proving DALI resolution correct and complete w.r.t. this
declarative semantics,
by means of an adaptation of standard proofs. 

It is important to notice that we have defined the semantics of our new language by 
modifying the program, while leaving the semantic approach unchanged. 
In this way, we keep all the useful properties of 
the Horn--clause language, thus we are still able to exploit
all the technical machinery related to it 
(such as methods for program analysis and optimization,
abstract interpretation, partial evaluation, debugging, etc.) which remains 
applicable on $P'$.

If we want to be more precise in our ``snapshot'' of the behavior of
agent $P$, we can simulate a resolution strategy $R\s$ by taking a program
$P'(R) \subseteq P'$, where $P'(R)$ differs form $P'$ in that 
only the clauses that $R\s$ should select are left, while the others are 
cancelled. Of course however, several $P'(R)$ can be obtained from $P'$. 

We can extend this semantic approach by defining the following procedure
for modeling evolutions of a multi--agent system:
\begin{enumerate}
\item
Given $n$ logic programs (agents) $P_1,\ldots,P_n$,
build ${P_1}',\ldots,{P_n}'$.
\item
Given computation rule $R\s$,
from ${P_1}',\ldots,{P_n}'$, build ${P_1}'(R),\ldots,{P_n}'(R)$.
\item
Compute the least Herbrand models of ${P_1}'(R),\ldots,{P_n}'(R)$,
say ${M_1},\ldots,{M_n}$.
\item
Taken as starting point the set of events contained in 
${M_1},\ldots,{M_n}$, 
change the unit clauses of ${P_1}'(R),\ldots,{P_n}'(R)$ accordingly,
and go back to step 3.
\end{enumerate}

In this way, we obtain possible {\em evolutions} of our agent system.
In order to study these evolutions, we can for instance pick up
the suggestion by \cite{giunchiglia:mca}, of adopting well--established 
techniques from the field of model--checking.
\section{Concluding Remarks }
\label{concl}
An agent as defined above, is completely characterized by its input channels, 
resolution strategy and Horn--clause theory.
About the input channels, of course we recognize the need, 
as emphasized by \cite{genesereth:ag},
of an underlying ``transducer'' integrated in the implementation, 
that converts all the incoming ``stimuli''
into a form that is intelligible to the agent and inserts them into the right channel.

A lot of implementation issues have been left open in this paper. An important
point is that of timely response to events, possibly guided by directives 
specifying time constraints. These constraints should influence the resolution
strategy. It is not clear however how one could prove that a DALI program
satisfies given real--time requirements.

Metareasoning is important for agents:
as remarked in \cite{martelliMAS}, an agent can reason on the basis
of meta--goals, which possibly require sophisticated strategies to be achieved.
Then, it would be interesting to take as basis 
for DALI an enhanced Horn--clause language
with self--reference and reflection like Reflective Prolog (\cite{CL:rp} which is the
basic language  for \cite{CDL:meta92} and  \cite{pier:jelia98}). 
On the lines of \cite{ROK} and \cite{CDL:meta92}, 
metalevel information could also be used to construct agent interfaces,
and to record information about heterogeneous information sources, in view of the
integration of heterogeneous information systems.

It is important to notice that we have not considered negation.
Negation is instead really important from the expressive and practical
point of view, at least since it allows exceptions to be stated to general reactive rules.
In the following example, where we assume to have negation
as failure, we define a situation where birds can fly for trying to 
avoid a predator (while the other animals run away), except for abnormal birds,
like penguins or road--runners. 

\begin{example}
\begin{eqnarray*}
Animal & & \\
\I{predator\_Attacks$_E$} &  \DE  & \I{try\_to\_Escape}.\\ 
\I{try\_to\_Escape} &  \IF  & \I{fly$_A$}.\\
\I{try\_to\_Escape} &  \IF  & \I{run$_A$}.\\
\I{fly$_A$} &  \DA  & \I{bird}, \no \I{abnormal}.
\end{eqnarray*}
\end{example}

Negation adds further complication to both declarative and procedural semantics. 
Nevertheless, adding negation to DALI is a main topic of future research.

Since this is work in progress, a suitable comparison with related work
is missing. We mean to add the comparison in the final version of the paper.

\end{sloppypar}
\footnotesize
\newpage
%
%

%
\end{document}